\documentclass[10pt,twocolumn,letterpaper]{article}

\usepackage{cvpr}              % To produce the CAMERA-READY version
\usepackage{graphicx}
\usepackage{amsmath}
\usepackage{amssymb}
\usepackage{booktabs}
\usepackage{blindtext}
\usepackage{caption}

\usepackage{multirow}
\usepackage{bbm}

\usepackage[pagebackref,breaklinks,colorlinks]{hyperref}

\usepackage[capitalize]{cleveref}
\crefname{section}{Sec.}{Secs.}
\Crefname{section}{Section}{Sections}
\Crefname{table}{Table}{Tables}
\crefname{table}{Tab.}{Tabs.}

\def\cvprPaperID{7998} % *** Enter the CVPR Paper ID here
\def\confName{CVPR}
\def\confYear{2023}

\title{A Hierarchical Representation Network for Accurate and Detailed Face Reconstruction from In-The-Wild Images}

\author{Biwen Lei, Jianqiang Ren, Mengyang Feng, Miaomiao Cui, Xuansong Xie\\
DAMO Academy, Alibaba Group\\
{\tt\small \{biwen.lbw, jianqiang.rjq, mengyang.fmy, miaomiao.cmm\}@alibaba-inc.com, }\\
{\tt\small xingtong.xxs@taobao.com}
}
\begin{document}
% \twocolumn[{
% \renewcommand\twocolumn[1][]{#1}
% \maketitle
 
% \begin{figure*}[t]
%    \setlength{\belowcaptionskip}{-0.3cm}
%    \setlength{\abovecaptionskip}{0.1cm}
% \centerline{\includegraphics[scale=0.55]{figures/first_page_figure.pdf}}
% \caption{Example of different levels of facial details and the hierarchical representation.}
% \label{fig:first}
% \end{figure*}

\twocolumn[{
\renewcommand\twocolumn[1][]{#1}
\maketitle
\vspace{-30pt}
\begin{center}
    \centering
    \captionsetup{type=figure}
    \includegraphics[scale=0.49]{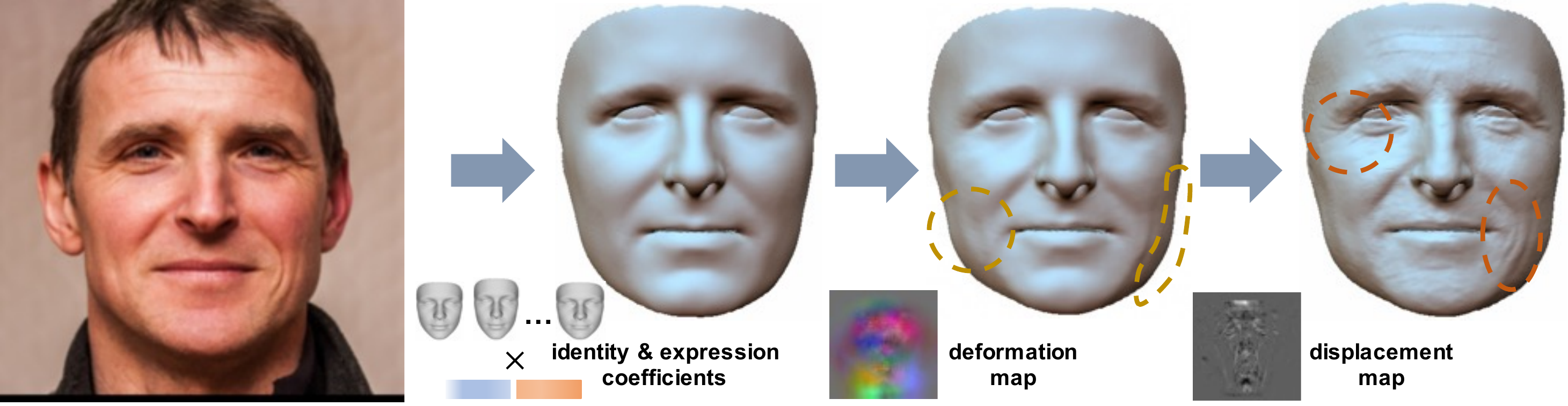}
    % \captionof{figure}{Example of different levels of facial details and the hierarchical representation.}
    \captionof{figure}{Example of low-frequency geometry, MF, HF facial details and the hierarchical representation.}
    \label{fig: first_page}
\end{center}
}]

% \teaser{
% \begin{center}
%     \centering
%     \captionsetup{type=figure}
%     \includegraphics[scale=0.50]{figures/first_page_figure.pdf}
%     % \captionof{figure}{Example of different levels of facial details and the hierarchical representation.}
%     \captionof{figure}{Example of low-frequency geometry, MF, HF facial details and the hierarchical representation.}
%     \label{fig: first_page}
% \end{center}
% }
% \maketitle

%-------------------------------------------------------------------------

%%%%%%%%% ABSTRACT
\begin{abstract}
\vspace{-10pt}

 Limited by the nature of the low-dimensional representational capacity of 3DMM, most of the 3DMM-based face reconstruction (FR) methods fail to recover high-frequency facial details, such as wrinkles, dimples, etc. Some attempt to solve the problem by introducing detail maps or non-linear operations, however, the results are still not vivid. To this end, we in this paper present a novel hierarchical representation network (HRN) to achieve accurate and detailed face reconstruction from a single image. Specifically, we implement the geometry disentanglement and introduce the hierarchical representation to fulfill detailed face modeling. Meanwhile, 3D priors of facial details are incorporated to enhance the accuracy and authenticity of the reconstruction results. We also propose a de-retouching module to achieve better decoupling of the geometry and appearance. It is noteworthy that our framework can be extended to a multi-view fashion by considering detail consistency of different views. Extensive experiments on two single-view and two multi-view FR benchmarks demonstrate that our method outperforms the existing methods in both reconstruction accuracy and visual effects. Finally, we introduce a high-quality 3D face dataset FaceHD-100 to boost the research of high-fidelity face reconstruction. The project homepage is at \href{https://younglbw.github.io/HRN-homepage/}{https://younglbw.github.io/HRN-homepage/}.

%Limited by the intrinsic low-dimentional representation, most of the 3DMM based methods fail to recover high-frequency details, such as wrinkles, dimples, etc. Some works attempt to solve the problem by leveraging detail maps or non-linear operations, but the results are still in lack of fineness and accuracy. In this paper, we propose a novel hierarchical representation network (HRN) to achieve accurate and detailed face reconstruction (FR) from single image. Firstly, we introduce the hierarchical representation to fulfill facial geometry disentanglement and modeling. Then, the 3D priors of details are incorporated to enhance the realism and accuracy of the reconstruction results. Besides, we propose a de-retouching module to achieve better decoupling of the geometry and appearance. Thanks to the powerful expressive ability brought by the hierarchical representation, we successfully extend the framework to a multi-view fashion by adding detail consistency between different views. Extensive experiments on two single-view and two multi-view FR benchmarks reveal that our method outperforms the existing methods in both reconstruction accuracy and detail effects. Moreover, we introduce a high quality 3D face dataset FaceHD-100 to boost the research of high fidelity face reconstruction.

\end{abstract}

%%%%%%%%% BODY TEXT

\vspace{-17pt}

\section{Introduction}

\vspace{-1pt}

High-fidelity 3D face reconstruction finds a wide range of applications in many scenarios, such as AR/VR, medical treatment, film production, etc. While extensive works already achieved excellent reconstruction performance using specialized hardware like LightStage\cite{2009The, cao2018sparse, ma2007rapid}, estimating highly detailed face models from single or sparse-view images is still a challenging 
problem. Based on 3DMM\cite{blanz1999morphable}, a statistical model learned from a collection of face scans, many works\cite{2019GANFIT, 2020AvatarMe, 2021Towards, 2022Fast} attempt to reconstruct the 3D face from a single image and achieve impressive results. However, limited by the nature of the low dimensional representational ability of the 3DMM, these methods can not recover the detailed facial geometry.

Recently, some methods\cite{chen2019photo,2018CNN,2017Learning} devote to capturing high-frequency facial details such as wrinkles by predicting a displacement map. They achieve realistic results, however, fail to model the mid-frequency details, such as the detailed contour of the jaw, cheeck, etc. To this end, some works try to capture the overall details by introducing latent encoding of details\cite{feng2021learning} or non-linear operations\cite{2019Towards, feng2018joint}. Nevertheless, it is hard to make a trade-off when handling the mid- and high-frequency details simultaneously. 
%facing the trade-off between the different regularizations for the mid-frequency and high-frequency details, these methods can hardly handle both levels of details well simultaneously, leading to dissatisfying results.
Another challenge is how to obtain accurate shapes and detailed 3D facial priors considering multifarious lightings and skins for different images.
%, due to the deficiency of 3D prior to the real facial details, most of the existing methods fail to obtain accurate shapes or be vulnerable to various lighting and skin texture. 
\cite{chen2019photo, 2015Real} resort to the wrinkle statistics computed from 3D face scans to fulfill realistic high-frequency details, but still fail to model the mid-frequency details.

\begin{figure*}[t]
   \setlength{\belowcaptionskip}{-0.3cm}
   \setlength{\abovecaptionskip}{0.1cm}
\centerline{\includegraphics[scale=0.35]{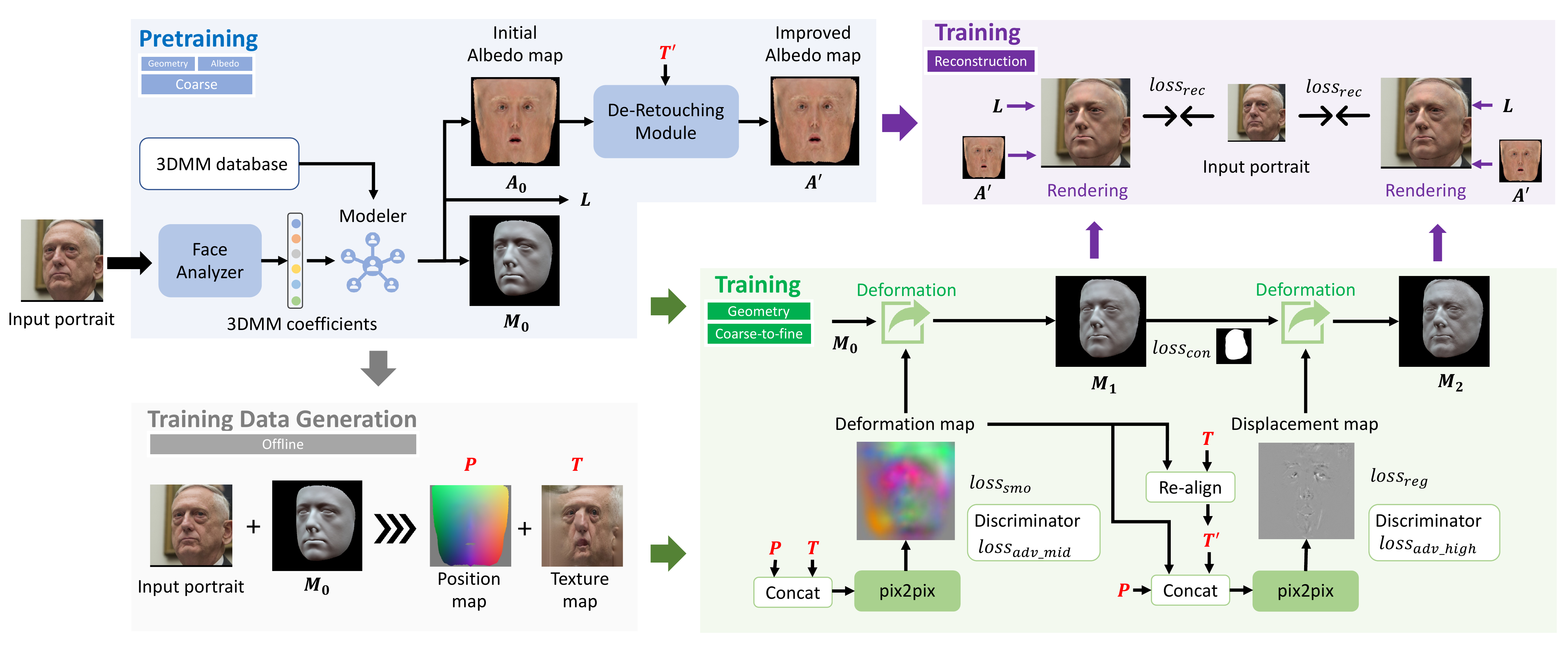}}
\caption{Overview of the proposed hierarchical representation network (HRN).}
\label{fig:framework}
\end{figure*}

Based on the observations above, we introduce a hierarchical representation network (HRN) for accurate and detailed face reconstruction from single image, as shown in Fig.~\ref{fig:framework}.
%, as shown in Fig.~\ref{fig:framework}. 
Firstly, we decouple the facial geometry into low-frequency geometry, mid-frequency (MF) details, and high-frequency (HF) details. Then, in a hierarchical fashion, we model these parts with face-wise blendshape coefficients, vertex-wise deformation map, and pixel-wise displacement map, respectively (shown in Fig.~\ref{fig: first_page}). Concretely, we employ two image translation networks\cite{2016Image} to estimate the corresponding detail maps (deformation and displacement map), and further employ them to generate the detailed face model in a coarse-to-fine manner. Moreover, we introduce the 3D priors of MF and HF details by fitting face scans with our hierarchical representation to facilitate accurate and faithful modeling. Inspired by\cite{lei2022abpn}, we propose a de-retouching module to adaptively refine the base texture to overcome the ambiguities between skin blemishes and illuminations. Extensive experiments show that our method outperforms the existing methods on two large-scale benchmarks, exhibiting excellent performance in terms of detail capturing and accurate shape modeling. Thanks to the detail disentanglement strategy and the guidance of detail priors, we extend HRN to a multi-view fashion and achieve accurate FR from only a few views.
%, making full use of the clues indicating the facial details from the 2D images. 
% We leverage a canonical space to describe the intrinsic shape and a view-dependent space to model the different poses, and lighting among the different views. Then an optimization-based framework is proposed to achieve accurate face reconstruction from sparse-view images. 
Finally, to boost the research of sparse-view and high-fidelity FR, we introduce a high-quality 3D face dataset named FaceHD-100.

Our main contributions in this work are as follows: \\
\textbf{(A)} We present a hierarchical modeling strategy and propose a novel framework HRN for single-view FR task. Our HRN produces accurate and highly detailed FR results and outperforms the existing state-of-the-art methods on two large-scale single-view FR benchmarks. \\
\textbf{(B)} We introduce detail priors to guide the faithful modeling of hierarchical details and design a de-retouching module to facilitate the decoupling of geometry and appearance. \\
\textbf{(C)} We extend HRN to a multi-view fashion to form MV-HRN, which enables accurate face modeling from sparse-view images and outperforms the existing methods on two large-scale multi-view FR benchmarks. \\
\textbf{(D)} To boost the research on sparse-view and high-fidelity FR tasks, we introduce a high-quality 3D face dataset FaceHD-100, containing 2,000 detailed 3D face models and corresponding high-definition multi-view images.

\section{Related Work}

\noindent {\bf Single-View Face Reconstruction.} Recovering 3D face from a single image is an ill-posed problem, but the advent of the 3D morphable model (3DMM)\cite{blanz1999morphable,romdhani2005estimating} has made it possible. The 3DMM provides strong prior knowledge and can represent complicated face geometry with low-dimensional coefficients. In this formulation, current methods can be categorized into either optimization-based\cite{blanz1999morphable,hu2017avatar,zhu2015high,blanz2004statistical,kemelmacher2011face} or learning-based\cite{deng2019accurate, tewari2017mofa,zielonka2022towards,ruan2021sadrnet}. Optimization-based approaches usually need costly analysis-by-synthesis processes and are sensitive to initialization, while learning-based methods directly train neural networks to regress the low-dimensional coefficients of 3DMM and recover 3D face through efficient forward inference. 

However, the original 3DMM models inherently lie in low-dimensional linear space and lack fine details. Many works\cite{chen2020self,tran2018nonlinear,sela2017unrestricted,feng2021learning,yang2020facescape} are proposed to overcome this limitation. Tran~\etal\cite{tran2018nonlinear} present a nonlinear 3DMM model and achieve more powerful representational abilities. Sela~\etal\cite{sela2017unrestricted} employ image-to-image network to generate pixel-based geometric representation for high quality reconstructions. In addition to static face geometry details, Feng~\etal\cite{feng2021learning} present an animatable displacement model to generate dynamic expression-depended wrinkles. Yang~\etal\cite{yang2020facescape} predict displacement maps via pix2pixHD network and combine them according to blendshape weights for dynamic details synthesis. Compared with these approaches, our method further introduces facial detail priors and can recover high fidelity facial details with hierarchical geometry representations.

\noindent {\bf Multi-View Face Reconstruction.}
Traditional multi-view stereo (MVS) methods\cite{furukawa2009accurate, bradley2010high,beeler2010high} are designed for 3D reconstruction given a set of multi-view images, but they heavily rely on the precision of camera calibration, and can hardly recover intact geometry in the sparse-view situation. To address these problems, many face-specialized multi-view reconstruction methods\cite{ichim2015dynamic,dou2018multi,piotraschke2016automated,ramon2019multi,wu2019mvf,bai2020deep,xiao2022detailed} are proposed. Ramon~\etal\cite{ramon2019multi} introduce siamese neural networks to extract relevant features from each view, and learn the 3D shape and the individual camera poses simultaneously. Wu~\etal\cite{wu2019mvf} exploit both 3DMM and
multi-view geometric constraints by estimating the alignment loss between multi-view inputs. Bai~\etal\cite{bai2020deep} leverage non-rigid multi-view stereo optimization to explicitly enforce multi-view appearance consistency, which is able to capture medium-level details. With the emergence of implicit 3D representation,  Xiao~\etal\cite{xiao2022detailed} propose to learn an implicit function to recover detailed geometry from calibrated multi-view images, but the implicit function learning is time-consuming which needs dozen of seconds and is sensitive to camera count and pose estimation. 

Rather than a specific design for multi-view inputs, our single-view model can be easily transferred to sparse-view face reconstruction task by adding hierarchical detail consistency between different views. Our method is robust to the calibration error of cameras thanks to the coarse-to-fine learning scheme.

\section{Methods}

\subsection{Overview}

In this paper, we propose a novel hierarchical representation network for accurate and detailed face reconstruction from single image. Fig.~\ref{fig:framework} illustrates the overview of our framework. We first employ 3DMM to predict a coarse mesh and albedo (blue area in Fig.~\ref{fig:framework}). Then we develop a hierarchical modeling (Sec.~\ref{sec:hierachical_modeling}) strategy to handle the complex facial details in a coarse-to-fine manner (green and purple area). To facilitate the accurate and faithful modeling of the hierarchical details, the 3D priors are incorporated through adversarial and semi-supervised learning (Sec.~\ref{sec:3D_priors}). Besides, we propose a de-retouching module (Sec.~\ref{sec:de-retouching}) to fulfill better decoupling of the geometry and appearance, alleviating the ambiguities between the various skin texture and illuminations. Moreover, we extend our framework to a multi-view fashion (Sec.~\ref{sec:MV-HRN}) and introduce a high quality 3D face dataset (Sec.~\ref{sec:dataset}) to boost the research on sparse-view face reconstruction. For simplification, we specify the related loss functions and training strategy in each section.

\subsection{Hierarchical Modeling} \label{sec:hierachical_modeling}

3DMM exhibit great performance in expressing various shape of the face, while the low-dimensional representation itself severely stem its learning on the details, leading to the imperfect alignment to the face or the over-smoothed results. Some methods extend 3DMM by introducing displacement maps to reconstruct some details such as wrinkles and bumps. However, focused on the high-frequency part, a simple displacement map still fails to handle some larger-scale details, such as the contours of the jaw and cheek. Based on such observation, we decouple the facial geometry into three components: (1) low-frequency part, which provides the coarse shape roughly aligned to the input face; (2) mid-frequency details, which describe the details of the contour and local shape relative to the low-frequency part; (3) and high-frequency details, such as wrinkles, micro bumps, etc. As shown in the Fig.~\ref{fig: first_page}, the scales decrease from the low-frequency part to the high-frequency detail, while the fineness increases in turn.

We design the hierarchical representation to model the above three components respectively. For the low-frequency part, we adopt the BFM as our base model and output the low-dimensional coefficients to fulfill a coarse reconstruction of the input face. Then we introduce a three-channel deformation map, which lies in the UV space and indicates the offset of each vertex relative to the coarse result. Worked as the representation of the mid-frequency details, the deformation map provides a flexible way to manipulate the geometries. We use the size of $64\times64$ to represent the deformation map to balance the fineness and smoothness of the mid-frequency details. For the high-frequency details, we employ the displacement map following\cite{feng2021learning}, which is a one-channel map ($256\times256$) denoting the geometry deformation along the direction of the normals. The displacement map is converted to detailed normals used in the rendering process in a pixel-wise manner to exhibit all the tiny details, breaking the limitation of the vertex density of the base model. Accordingly, we are enabled to describe an arbitrary complex face with these representations. 

As shown in Fig.~\ref{fig:framework}, given a portrait image $\boldsymbol{I}$, we firstly utilize a regression network as the face analyzer to predict the BFM coefficients and obtain the coarse aligned mesh $\boldsymbol{M_0}$ and albedo $\boldsymbol{A_0}$ using the corresponding basis from the 3DMM database. Combined $\boldsymbol{I}$ and $\boldsymbol{M_0}$, we are able to acquire the inpainted texture $\boldsymbol{T}$ in UV space by applying the differentiable rendering with a coarse-to-fine strategy. And we concatenate the position map $\boldsymbol{P}$ and $\boldsymbol{T}$ as the input of the following modules for hierarchical details learning. We adopt two pix2pix\cite{2016Image} networks to synthesize the deformation map and displacement map in sequence.
% which are further applied to $M_0$ to generate the detailed mesh $M_1$ with MF details and $M_2$ with hierarchical details in a coarse-to-fine manner. 
Note that, considering the deformation map will change the facial geometry and lead to the misalignment between ${\boldsymbol{T}}$ and the deformed mesh, we generate the realigned texture $\boldsymbol{T^{'}}$ as the input of second pix2pix network by projecting the three-channel deformation map to the 2D space and transform it to a reversal flow ${\boldsymbol{F}}$ to re-align ${\boldsymbol{T}}$. Taking advantage of the abundant details from ${\boldsymbol{T}}$ and the pixel-wise learning strategy, we manage to obtain the accurate detail maps which are further employed in a coarse-to-fine manner to generate the detailed face mesh ${\boldsymbol{M_1}}$ and ${\boldsymbol{M_2}}$. Finally, combined with the lighting ${\boldsymbol{L}}$ and the refined albedo generated from the de-retouching module (Sec.~\ref{sec:de-retouching}), we accomplish detailed face reconstruction from the single image. 

Overall, the framework is trained in a self-supervised manner guided with 3D detail priors learned from face scans (Sec.~\ref{sec:3D_priors}). To reduce the training complexity, we adopt the pre-trained encoder and MLP from the\cite{deng2019accurate} as the face analyzer to predict the coefficients and generate the corresponding $\boldsymbol{P}$ and $\boldsymbol{T}$ for the following details learning. The two image translation networks are trained jointly and the related loss functions are composed of three components:

\noindent {\bf Reconstruction Loss.} The reconstruction loss is calculated between the rendered face and the input face and consists of the photometric loss $L_{photo}$ , perception-level loss $L_{per}$ and landmark loss $L_{lan}$ following\cite{deng2019accurate}. Thanks to the delighted albedo and the illumination system of 3DMM, the photometric loss will enforce the deformation of the facial geometry to fit the various shadows and highlight areas of the input face. It is crucial that we apply the reconstruction loss on both images rendered from the $\boldsymbol{M_1}$ and $\boldsymbol{M_2}$, which benefit the disentanglement of the MF and HF details.

\noindent {\bf Details Loss.} We apply the total variation loss $L_{tv}$\cite{johnson2016perceptual} to encourage the smoothness of the deformation map, and use the L1 regularization loss $L_{reg}$ to limit the scale of the displacement map.

\begin{figure}[t]
  \centering
  \resizebox{0.98\linewidth}{!}{
   \includegraphics{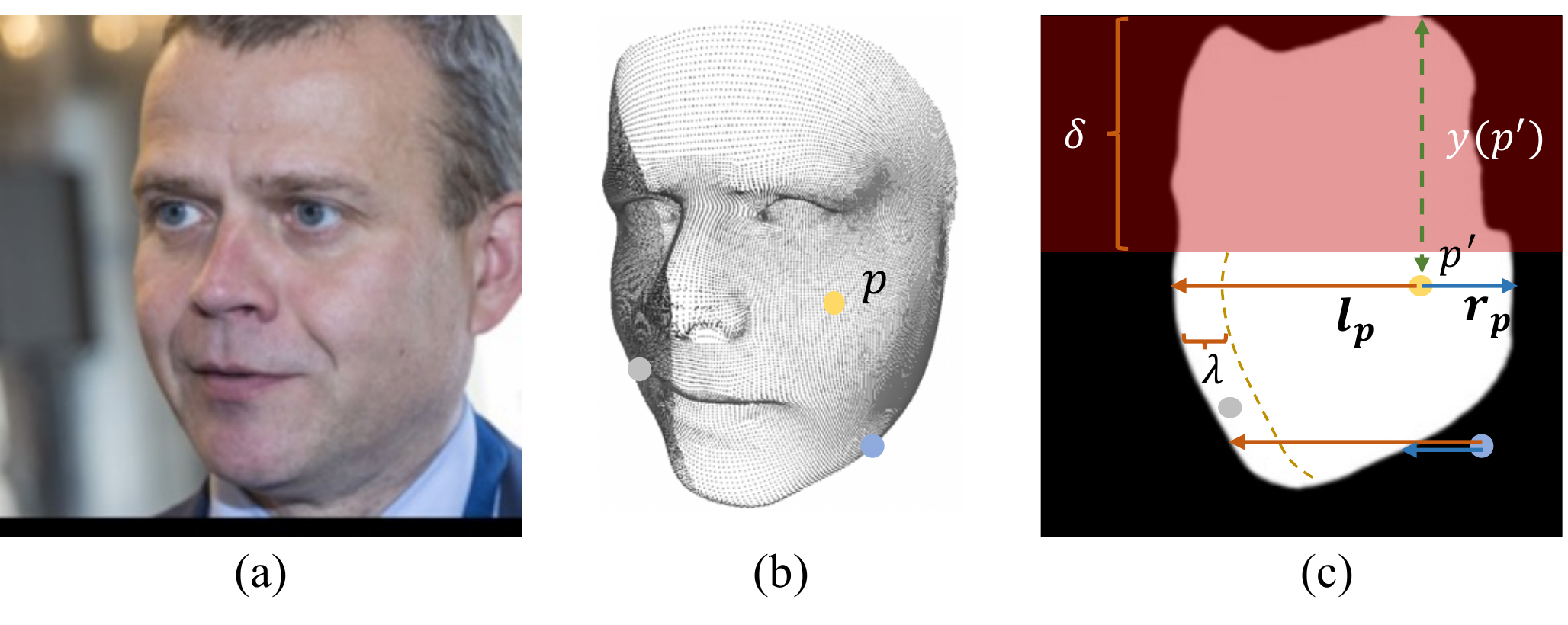} }
     \vspace{-10pt}
  \caption{The details of the proposed contour loss. (a) input image, (b) the projected vertices, (c) the predicted face mask.}
  \label{fig: contour_loss}
  % \vspace{-10pt}
\end{figure}

\noindent {\bf Contour-aware Loss.} We propose a novel contour-aware loss $L_{con}$ to fulfill accurate modeling of the face contour. The $L_{con}$ works on $\boldsymbol{M_1}$ and aims to pull the vertices of edge to align the face contour. As shown in Fig.~\ref{fig: contour_loss}, we firstly project vertices of $\boldsymbol{M_1}$ into the image space. Then we predict the face mask $\boldsymbol{M_{face}}$ using the pre-trained face matting network\cite{liu2020boosting} and implement post process to obtain the left side and right side points for each row. Given a vertex $p$ and the corresponding projected point $p'$ on $\boldsymbol{M_{face}}$, we obtain the vector $\boldsymbol{l_p}$ and $\boldsymbol{r_p}$ (from $p'$ to the edge points in the horizontal direction). Then $L_{con}$ can be describe as:
\begin{equation}\label{eq:contour_loss}
    {L_{con}} = \frac{1}{N_p} {\sum_{p \in M_1} (f(p) \mathbbm{1}[y(p')>\delta])},
\end{equation}
\begin{equation}\label{eq:dist_func}
    {f(p)} = \lvert h(\frac{\boldsymbol{l_p} \cdot \boldsymbol{r_p}}{max(\lvert \lvert \boldsymbol{l_p} \rvert \rvert, \lvert \lvert \boldsymbol{r_p} \rvert \rvert)} + \lambda) - \lambda \rvert,
\end{equation}
where $h$ is the ReLU function, and $\lambda$ denotes a soft margin relative to the face contour ($\lambda=0.01$ as default), $\mathbbm{1}[y(p')>\delta]$ indicates whether $p'$ is on the lower part of the image ($\delta=100$ as default). As we can see, $L_{con}$ punish the vertices outside the soft margin (such as the blue and gray points in Fig.~\ref{fig: contour_loss}) of the face and pull them to the face contour, while keeping the vertices inside the face intact. Combined with $L_{tv}$ of the deformation map, $L_{con}$ will avoid the unsmooth effect near the face contour. Note that we only focus on the lower part of the face contour to avoid the distraction of the hair.
Compared to the common segmentation loss, $L_{con}$ gives a more straightforward direction for optimizing the face contour and is easier for training. We conduct an ablation study to reveal the effectiveness of $L_{con}$ in Sec.~\ref{sec:ablation}.

\subsection{3D Priors of Facial Details} \label{sec:3D_priors}

% We are capable of learning the facial details from the solely 2D input image as well by applying the reconstruction loss (Sec.~\ref{sec:hierachical_modeling}). However, it is a highly ill-posed problem and has innumerable possible solutions that is unreal. Strong regularization can narrow the solution space, but it also severely downgrade the accuracy and fidelity of details. 
Although facial details can be roughly learned from single image using the reconstruction loss (Sec.~\ref{sec:hierachical_modeling}), it suffers from unreality and ambiguousness due to its ill-posed essence. Adding additional regularization may help to narrow the solution space, but also lead to severe degradation in detail accuracy and fidelity.
 
\begin{figure}[t]
  \centering
  \resizebox{0.98\linewidth}{!}{
   \includegraphics{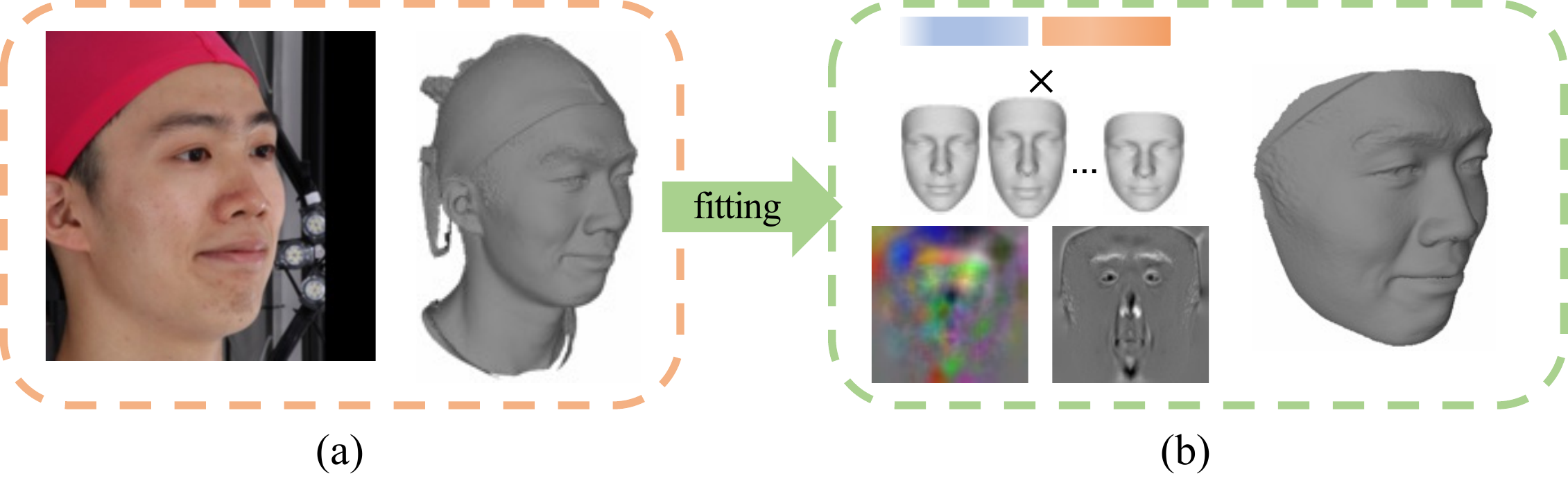} }
     \vspace{-5pt}
  \caption{An example of the 3D priors of facial details. (a) the raw image and scan, (b) the hierarchical representation and the corresponding fitted mesh.}
  \label{fig: 3D_priors}
  \vspace{-5pt}
\end{figure}

To address this problem, we exploit the 3D priors of facial details derived from face scans and corresponding multi-view images in our framework. Firstly, given an raw image and its corresponding scan, we transform the raw scan to align to the image in BFM space (the details can be found in the supplemental files). Then we can obtain the ground-truth deformation map and displacement map for each image by fitting the image and scan using the loss functions mentioned in Sec.~\ref{sec:hierachical_modeling} with additional supervision on vertices distance following\cite{amberg2007optimal} . Thanks to the powerful hierarchical representation, the details of scans can be accurately captured. See Fig.~\ref{fig: 3D_priors} for example. 

% \begin{figure}[t]
%   \centering
%   \resizebox{0.98\linewidth}{!}{
%    \includegraphics{figures/3D_priors.pdf} }
%      \vspace{-10pt}
%   \caption{The capturing process of the 3D priors of facial details. (a) raw scans, (b) transformed scans that is aligned to the BFM space, (c) the captured hierarchical representation.}
%   \label{fig: 3D_priors}
%   \vspace{-10pt}
% \end{figure} 

We take advantage of the 3D priors of details on two aspects. On the one hand, we develop two discriminators and use the adversarial loss\cite{isola2017image} $L_{adv\_mid}$ and $L_{adv\_high}$ to supervise the domain distribution of the deformation map and displacement map. On the other hand, we acquire the paired data as mentioned above from 3D scans to conduct supervised learning to guide the self-supervised learning in Sec.~\ref{sec:hierachical_modeling}. Specifically, we supplement the L1 loss $L_{mid}$ and $L_{high}$ for the predicted deformation map and displacement map respectively. Note that a mask is used in training to remove the distraction of eyes and hair area from scans.

\subsection{De-Retouching Module} \label{sec:de-retouching}

A face image is the result of a combination of geometry, lighting, and face albedo. Prior works assume that the face albedo is smooth and model it with the low-frequency albedo from 3DMM. However, the actual skin texture is full of high-frequency details such as moles, scars, freckles, and other blemishes, which bring ambiguities to the geometry details learning especially in the single view FR task. Inspired by the\cite{lei2022abpn}, we propose a de-retouching module (DRM), which aims to generate the face albedo with high-frequency details and facilitate more precise decoupling of geometry and appearance.

\begin{figure}[t]
  \centering
  \resizebox{0.98\linewidth}{!}{
   \includegraphics{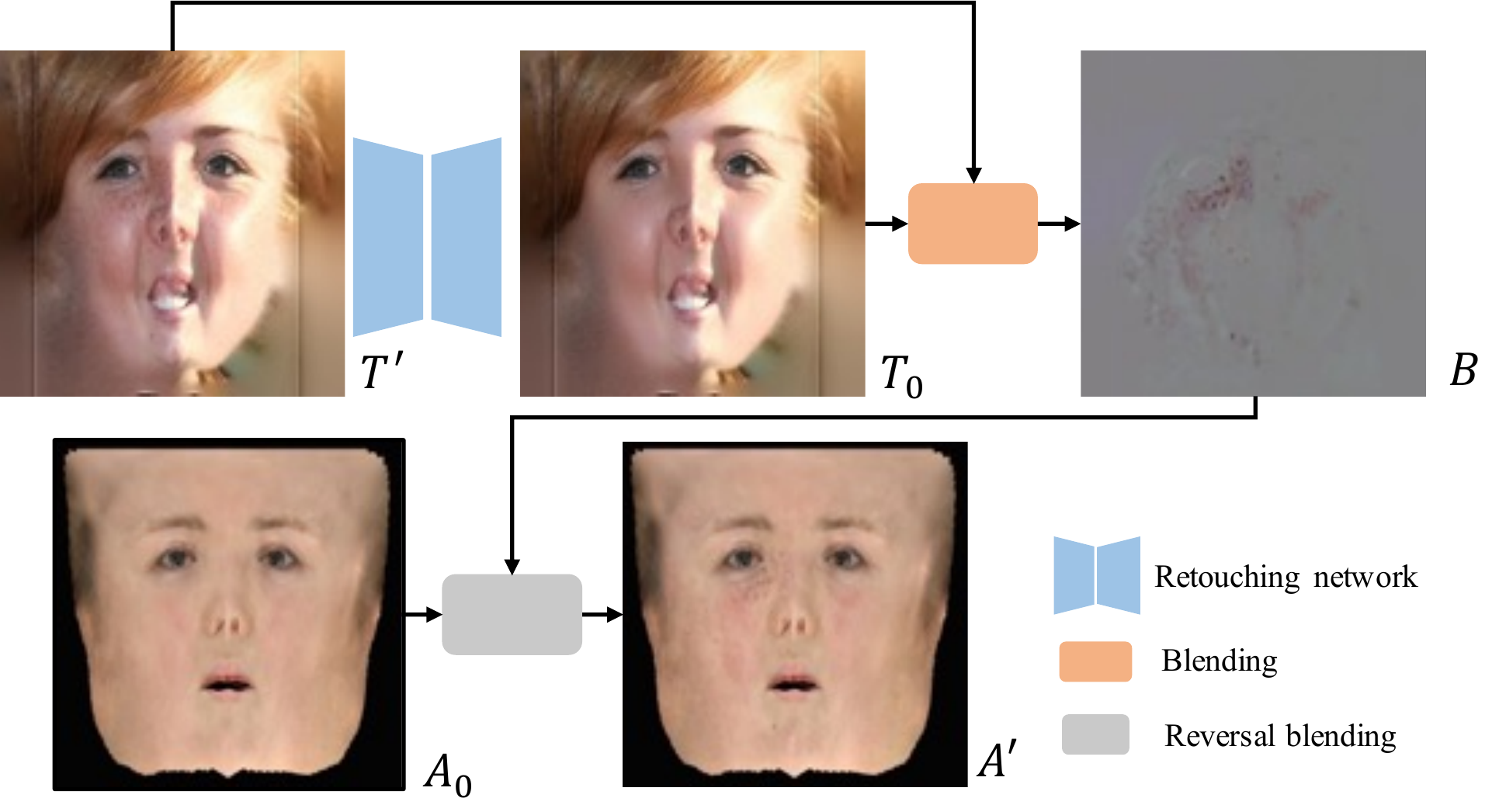} }
     \vspace{-10pt}
  \caption{The details of the de-retouching module.}
  \label{fig: de-retouching}
  \vspace{-5pt}
\end{figure} 

\begin{figure}[t]
  \centering
  \resizebox{0.98\linewidth}{!}{
   \includegraphics{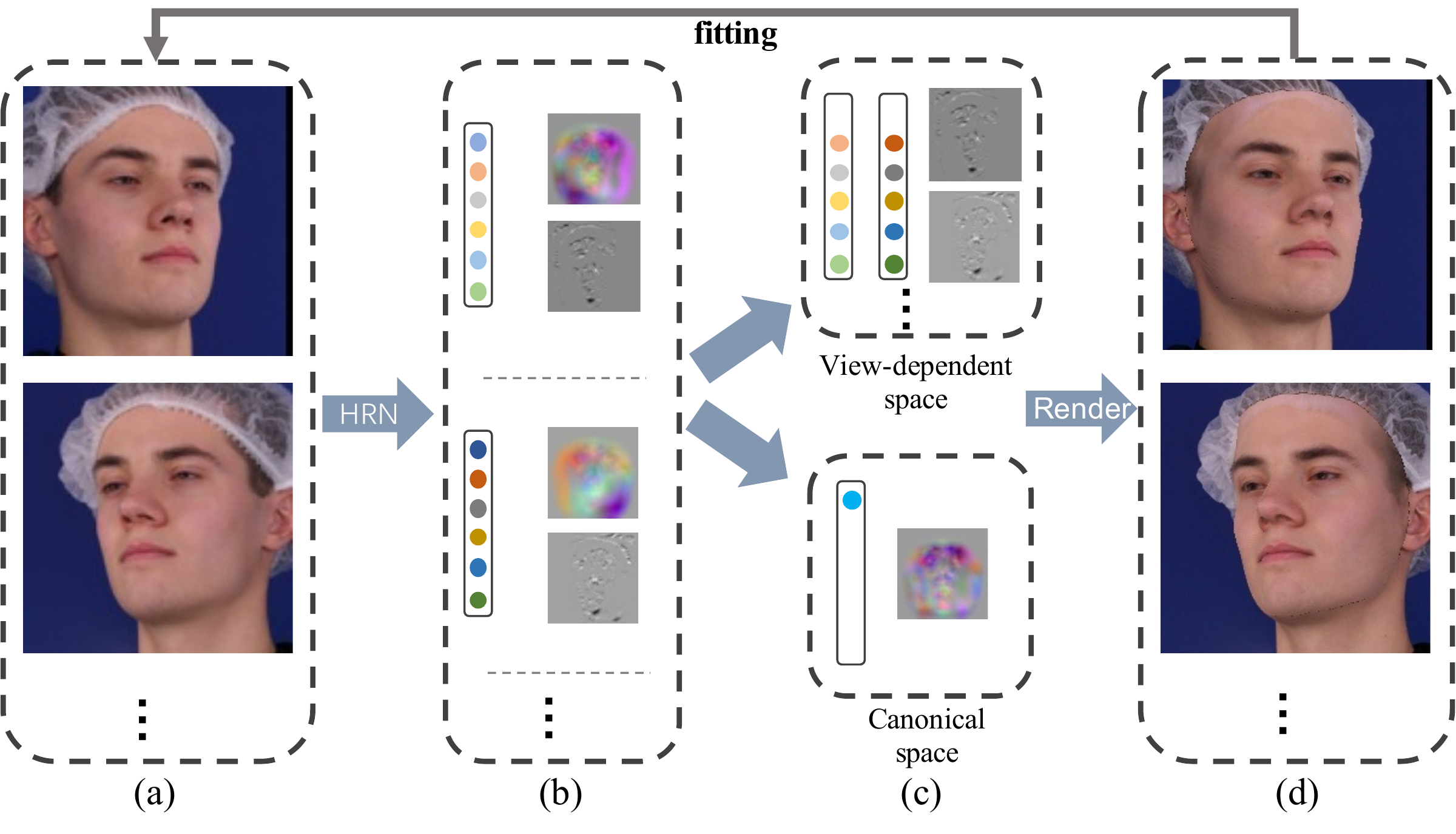} }
     \vspace{-5pt}
  \caption{The pipeline of the proposed MV-HRN.}
  \label{fig: MV-HRN}
  \vspace{-10pt}
\end{figure} 

We collect 10, 000 face images from FFHQ\cite{karras2019style}, and hire a team of professional image editors to process the images, with the goal of removing the skin blemishes and other texture details while maintaining the shape-related content such as wrinkles, bumps, etc. Then we transform the paired images into the texture maps in UV space by applying the process specified in Sec.~\ref{sec:hierachical_modeling} and train an image translation network $G$ to fulfill skin retouching. Given the re-aligned texture $\boldsymbol{T^{'}}$, we firstly employ $G$ to remove its texture details and get $\boldsymbol{T_0}$, as shown in Fig.~\ref{fig: de-retouching}. We aim to bake the texture details into the coarse albedo $\boldsymbol{A_0}$ to obtain the improved albedo $\boldsymbol{A^{'}}$ for rendering. We make an assumption that the shading from $\boldsymbol{A_0}$ to $\boldsymbol{T_0}$ should be consistent with the one from $\boldsymbol{A^{'}}$ to $\boldsymbol{T^{'}}$, as: 

\begin{equation}\label{eq:deretouching_1}
    \boldsymbol{T_0} = \boldsymbol{A_0} \odot \boldsymbol{S},
\end{equation}
\begin{equation}\label{eq:deretouching_2}
    \boldsymbol{T^{'}} = \boldsymbol{A^{'}} \odot \boldsymbol{S},
\end{equation}
where $\boldsymbol{S}$ denotes the shading map, $\odot$ denotes element-wise matrix multiplication. Then we can solve the equations and obtain $A^{'}$ as:
\begin{equation}\label{eq:deretouching_3}
    \boldsymbol{A^{'}} = \boldsymbol{A_0} \odot \boldsymbol{B} \approx \boldsymbol{A_0} \odot \frac{\boldsymbol{T^{'}} + \phi (\boldsymbol{T_0})}{\boldsymbol{T_0} + \phi (\boldsymbol{T_0})},
\end{equation}
\begin{equation}\label{eq:deretouching_4}
    \phi (\boldsymbol{T_0}) = \frac{1}{\frac{\boldsymbol{T_0} \odot \boldsymbol{T_0} \odot \boldsymbol{T_0}}{\varepsilon} + \varepsilon},
\end{equation}
where $\phi (\boldsymbol{T_0})$ avoids the value explosion near 0 and $\varepsilon=1e-6$ as default. Compared to $\boldsymbol{A_0}$, the de-retouched albedo $\boldsymbol{A'}$ contains more high-frequency texture details, which alleviate the ambiguities between geometry and appearance, especially in single view FR task.

\subsection{MV-HRN} \label{sec:MV-HRN}

Thanks to the hierarchical modeling and the 3D priors guidance, we can easily adapt the HRN to a multi-view fashion to fulfill precise modeling of the global facial geometry with only a few views by adding the geometry consistency between different views. Fig.~\ref{fig: MV-HRN} shows the pipeline of MV-HRN. We assume that the low-frequency identity part and the mid-frequency details are consistent between different views, while the lighting, expression, and HF details should be view-dependent to overcome the disturbance. Therefore, we develop a canonical space, which contains the shared identity coefficient and deformation map that are initialized by averaging all the single-view results, to represent the shared intrinsic face shape. Then other BFM coefficients and the displacement map of each view are utilized to dependently model the pose, lighting, expression and high-frequency details. Then we apply the loss functions mentioned in Sec.~\ref{sec:hierachical_modeling} and Sec.~\ref{sec:3D_priors} to iteratively optimize all the coefficients and detail maps. Through the fitting process, the face shape is gradually restricted to a smaller and more accurate space under the supervision of different views. Extensive experiments show that MV-HRN achieves accurate reconstruction given only a few (2 $\sim$ 5) views of images in a short time (less than one minute).

\subsection{FaceHD-100 Dataset} \label{sec:dataset}
\begin{figure}[t]
  \centering
  \resizebox{0.98\linewidth}{!}{
   \includegraphics{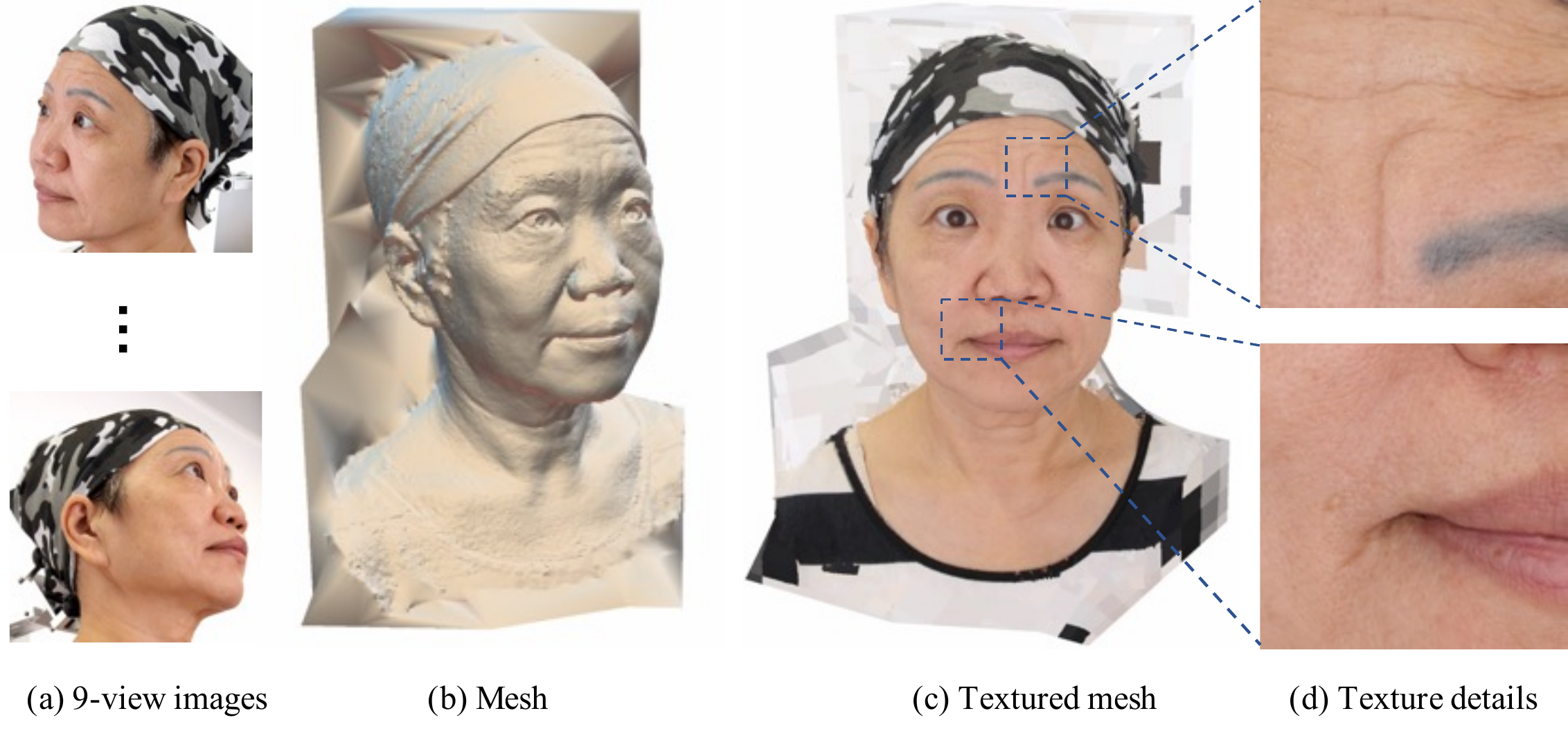} }
     \vspace{-10pt}
  \caption{An example from the FaceHD-100 dataset.}
  \label{fig: dataset}
  \vspace{-10pt}
\end{figure}

To boost the research of face reconstruction from sparse-view images, we introduce a high-quality 3D face dataset FaceHD-100, which consists of 2,000 high-definition 3D mesh and corresponding multi-view images from 100 subjects. The data is captured by a multi-view 3D reconstruction system, which is composed of 9 DSLR cameras and 8 LED lights. The 9 cameras are evenly distributed in front of and to the side of the face, and each provides 8K images for geometry and texture reconstruction. The capturing subjects include 50 males and 50 females, and mostly are from Asia. The ages of these subjects are normally distributed from 16 to 70 years old. For each person, we follow\cite{triplegangers} and ask them to perform 20 expressions including the neutral expression for capturing. Fig.~\ref{fig: dataset} gives an example of FaceHD-100, which shows the high quality of the reconstructed geometry and texture.

\section{Experiments}
\subsection{Implementation Details}
\noindent {\bf Training Data.} The training data of the proposed model is composed of two parts: 2D in-the-wild images and 3D face scans with corresponding multi-view images. For the former, we collect in-the-wild face images from multiple sources following\cite{deng2019accurate}. For the latter, the data is collected from FaceScape\cite{zhu2021facescape}, ESRC\cite{ESRC} and FaceHD-100. To be specific, we split 359 samples of FaceScape into training (309 subjects) and testing sets (50 subjects), considering the balance of gender and age. The ESRC is also split in the same way as\cite{bai2020deep} and the whole FaceHD-100 dataset is used for training. In total, we collected $\sim$9K scans from nearly 500 subjects of different ethnicities. The majority of subjects have 3D scans for at least 8 different expressions. Then we process all the scans and corresponding multi-view images in the way shown in Fig.~\ref{fig: 3D_priors} to generate the ground-truth deformation maps and displacement maps for each image. In the end, we collected $\sim$260K in-the-wild images for self-supervised training and $\sim$150K lab images with corresponding ground-truth details map for supervised training. The input images are preprocessed following\cite{deng2019accurate}.

\noindent {\bf Training Strategy.} Firstly, we employ the pretrained R-Net from\cite{deng2019accurate} as our face analyzer to predict the 3DMM coefficients, position map and texture map as mentioned in Sec.~\ref{sec:hierachical_modeling} for the following training. We use the paired texture maps mentioned in Sec.~\ref{sec:de-retouching} to train the de-retouching module. Finally, we fix the parameters of the face analyzer and de-retouching module and train the whole network with the input of face images, position maps and texture maps. We train our model using the Adam optimizer with a batch size of 4, an initial learning rate of 1e-4, and 800K iterations. Note that the model is trained alternately with in-the-wild images and lab images in a self-supervised and supervised manner respectively. More details about the parameters and training setting are specified in supplementary materials.

\begin{figure}[t]
  \centering
  \resizebox{0.98\linewidth}{!}{
   \includegraphics{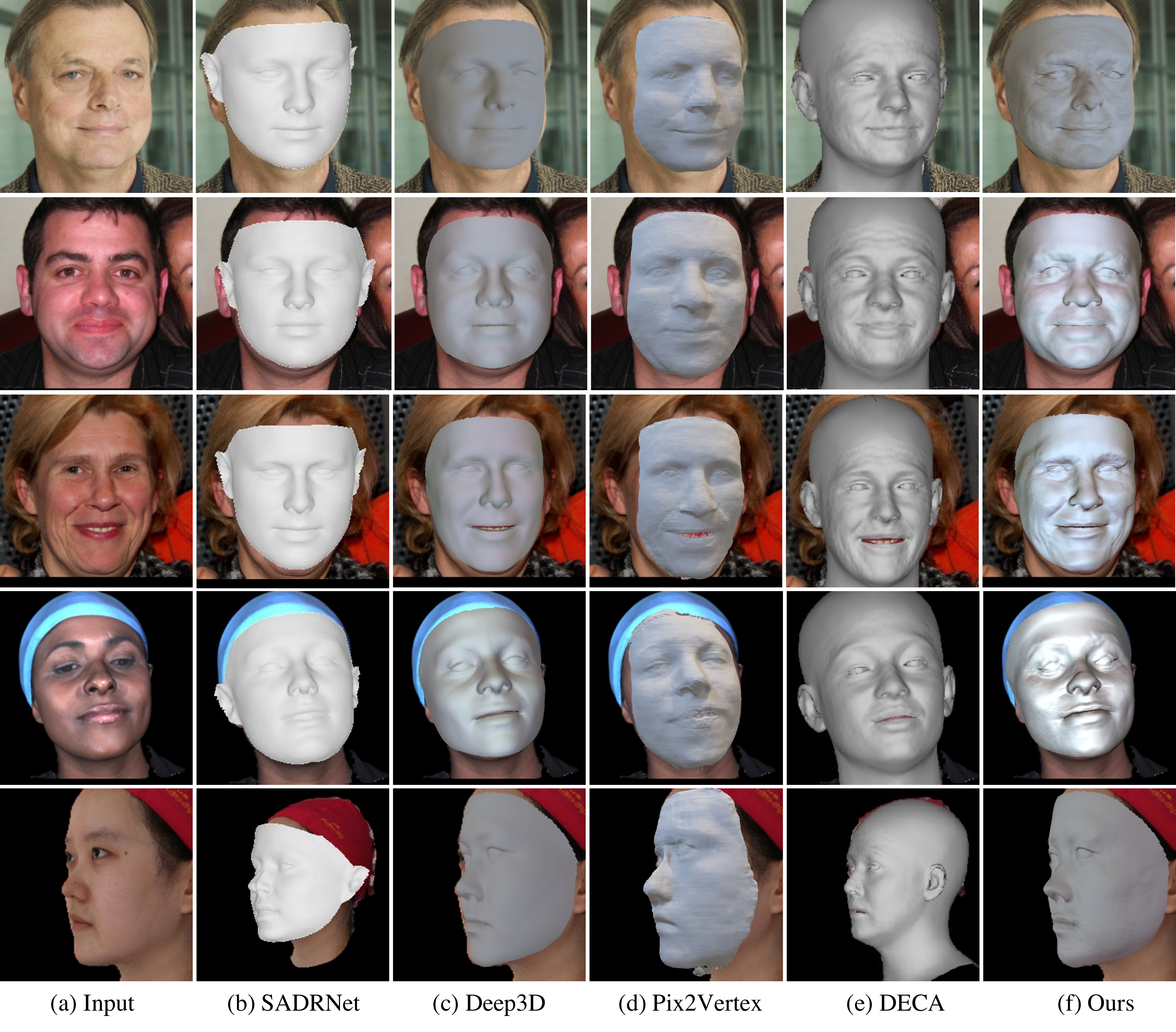} }
     \vspace{-10pt}
  \caption{The single-view qualitative comparison on FFHQ (first three rows), REALY (fourth row), and FaceScape (last row).}
  \label{fig: cmp}
  \vspace{-10pt}
\end{figure}

\begin{figure}[t]
  \centering
  \resizebox{0.98\linewidth}{!}{
   \includegraphics{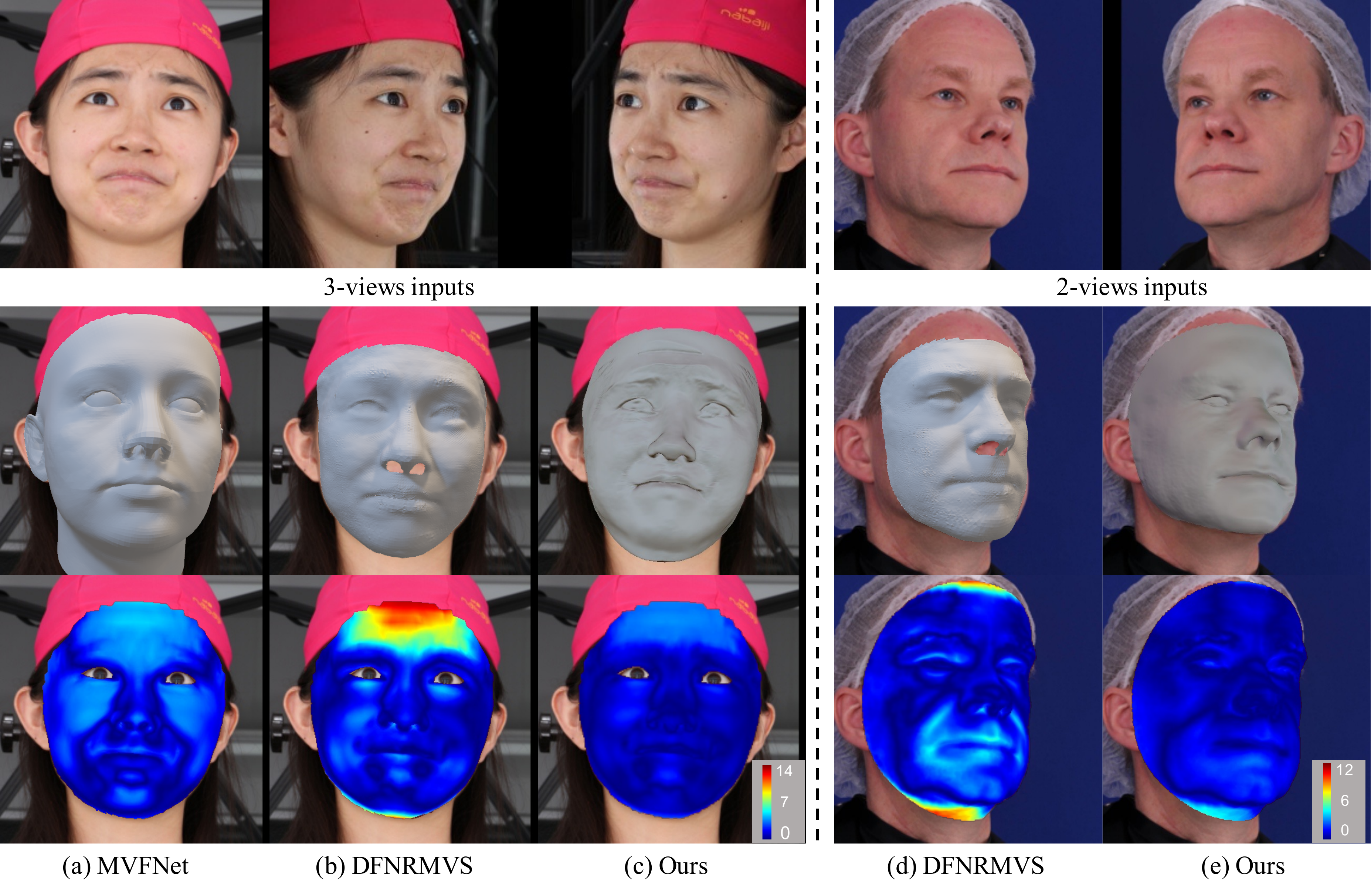} }
     \vspace{-5pt}
  \caption{The multi-view qualitative comparison on FaceScape and ESRC datasets.}
  \label{fig: multiview_cmp}
  \vspace{-15pt}
\end{figure}

\subsection{Qualitative Comparison}
We evaluate our model with four SOTA methods: SADRNet\cite{ruan2021sadrnet}, Deep3D\cite{deng2019accurate}, Pix2Vertex\cite{sela2017unrestricted}, and DECA\cite{feng2021learning} for single-view 3D FR task (we present comparisons with more methods\cite{chen2019photo,zhang2021learning} in the supplementary material). SADRNet tackles 3D dense face alignment and face reconstruction simultaneously with a self-aligned dual regression framework, and Deep3D leverages hybrid loss function to train CNN for 3DMM coefficients regression in a weakly-supervised manner. With the goal of adding more expressiveness and details, Pix2Vertex utilizes the Image-to-Image translation network to provide high-quality reconstructions under extreme expressions, while DECA presents an animatable displacement model to generate dynamic expression-depended face details. To make a fair comparison, we use their publicly released models and codes, and conduct experiments on FFHQ, REALY, and Facescape datasets.

The comparison results of the single-view reconstruction scenario are shown in Fig.~\ref{fig: cmp}. Since Deep3D only reconstructs faces using predicted 3DMM coefficients, the results are in low-dimensional space and lack high-frequency details. Although SADRNet tries to regress face shape deformation separately, the high-frequency details are still ignored in learning because of their minority. Pix2Vertex and DECA provide more fine details, such as blemishes and wrinkles. However, Pix2Vertex brings artifacts at the same time due to its unrestricted manner, and DECA cannot accurately recover face identities and expressions, resulting in similar wrinkles on the forehead among various faces. By contrast, our proposed method can produce high-fidelity 3D faces with expressive details, which are extremely consistent with the original input images.

In the multi-view scenario, we test the performance of DFNRMVS\cite{bai2020deep}, MVFNet\cite{wu2019mvf} and the proposed model given 3-view or 2-view images from FaceScape and ESRC datasets. As illustrated in Fig.~\ref{fig: multiview_cmp}, our model outperforms the other two methods in terms of fidelity, details and geometry accuracy, proving that our framework can be well generalized to both single-view and multi-view tasks.

\subsection{Quantitative Comparison}
Three public datasets are employed to quantitatively evaluate our method with several state-of-the-art approaches. Specifically, we choose FaceScape\cite{zhu2021facescape} dataset for both single-view and multi-view evaluation, and additionally use REALY\cite{REALY} and ESRC\cite{ESRC} datasets for single-view and multi-view respectively. To evaluate the geometry accuracy of single-view face reconstruction, we use Chamfer Distance (CD) and Mean Normal Error (MNE) on the FaceScape dataset following FaceScape benchmark\cite{zhu2021facescape}, while leveraging average Normalized Mean Square Error (NMSE) of different face regions on REALY dataset, which applies region-wise alignment and is more accurate for shape error computing. 

Table~\ref{tab:comp_sv} shows the quantitative comparison of single-view reconstruction. Our approach outperforms other methods on FaceScape-wild and REALY datasets, and achieves SOTA performance on the FaceScape-lab dataset. We find that our side-view metric is even better than the frontal-view on REALY dataset, we speculate that the slightly side-view images provide more information about mouth/node heights, which is beneficial to geometry prediction.

For multi-view reconstruction evaluation, as guided by NoW benchmark\cite{RingNet:CVPR:2019}, we firstly choose 7 face landmarks for each predicted mesh and then apply rigid alignment to ground truth mesh, and report their scan-to-mesh distances. Due to that MVFNet public model cannot handle 2-views images, we only use it in FaceScape dataset testing. The results in Table~\ref{tab:comp_mv} show that our approach performs better
against MVFNet and DFNRMVS with the lowest reconstruction errors. We also test the performance of our method on the MICC dataset\cite{bagdanov2011florence}, and the results are presented in supplementary materials.

\begin{table}[h]
 \centering
  \caption{Single-view quantitative comparison. REALY-F and REALY-S denote frontal-view and side-view reconstruction on REALY benchmark respectively.} % 
    \vspace{-5pt}
\footnotesize
\resizebox{1.00\linewidth}{!}{
\begin{tabular}{l|cc|cc|c|c}
    \toprule
\multirow{3}{*}{Methods} & \multicolumn{2}{c|}{FaceScape-wild} & \multicolumn{2}{c|}{FaceScape-lab} & REALY-F & REALY-S \\
    % & FS-wild   &FS-wild     & FS-lab  & FS-lab    & REALY-F    & REALY-S \\  

    % \toprule
 & CD  & MNE     & CD & MNE    & NMSE    & NMSE     \\
   & (mm)   & (rad)     & (mm)  & (rad)    & (mm)    & (mm)     \\
    \midrule
Deep3D     & 3.8      & 0.092          & 5.28     & 0.118   & 1.657     & 1.691          \\
MCGNet     & 3.22      & 0.077          & 4.00     & 0.093   & 1.774     & 1.787          \\
PRNet     & 3.47      & 0.123          & \textbf{3.56}     & 0.126   & 2.013     & 2.032          \\
SADRNet     & 7.12      & 0.123         & 6.75     & 0.133   & 1.913     & 1.958          \\
DECA     & 3.31      & 0.089          & 4.69     & 0.108   & 2.210     & 2.261          \\
3DDFA-V2     & 3.00      & 0.080         & 3.60     & 0.096    & 1.926     & 1.943          \\
Ours     & \textbf{2.91}      & \textbf{0.065}        & 3.67     & \textbf{0.087}   & \textbf{1.537}     & \textbf{1.468}          \\
    \midrule

\end{tabular}}
  \label{tab:comp_sv}
\end{table}
\vspace{-15pt}

\begin{table}[h]
 \centering
  \caption{Multi-view quantitative comparison. We only report MVFNet performance on FaceScape because its released model cannot process two-view inputs.} % 
    \vspace{-5pt}
\footnotesize
\resizebox{1.00\linewidth}{!}{
\begin{tabular}{l|ccc|ccc}
    \toprule
\multirow{3}{*}{Methods} & \multicolumn{3}{c|}{FaceScape (3 views)} & \multicolumn{3}{c}{ESRC (2 views)}   \\
    % & FS-wild   &FS-wild     & FS-lab  & FS-lab    & REALY-F    & REALY-S \\  

    % \toprule
   & Median & Mean & Std   & Median & Mean & Std  \\
   & (mm)   & (mm)     & (mm)  & (mm)    & (mm)    & (mm)     \\
    \midrule
MVFNet     &  1.76 & 2.12 & \textbf{1.66} & N.A. & N.A. & N.A.     \\
DFNRMVS     &  1.79 & 2.41 & 2.61 & 1.59 & 2.13 & 2.29     \\
Ours     &  \textbf{1.13} & \textbf{1.51} & 1.79 & \textbf{1.29} & \textbf{1.69} & \textbf{1.72}     \\
    \midrule

\end{tabular}}
  \label{tab:comp_mv}
\end{table}
\vspace{-15pt}

\subsection{Ablation Study} \label{sec:ablation}

\begin{figure}[t]
  \centering
  \resizebox{0.98\linewidth}{!}{
   \includegraphics{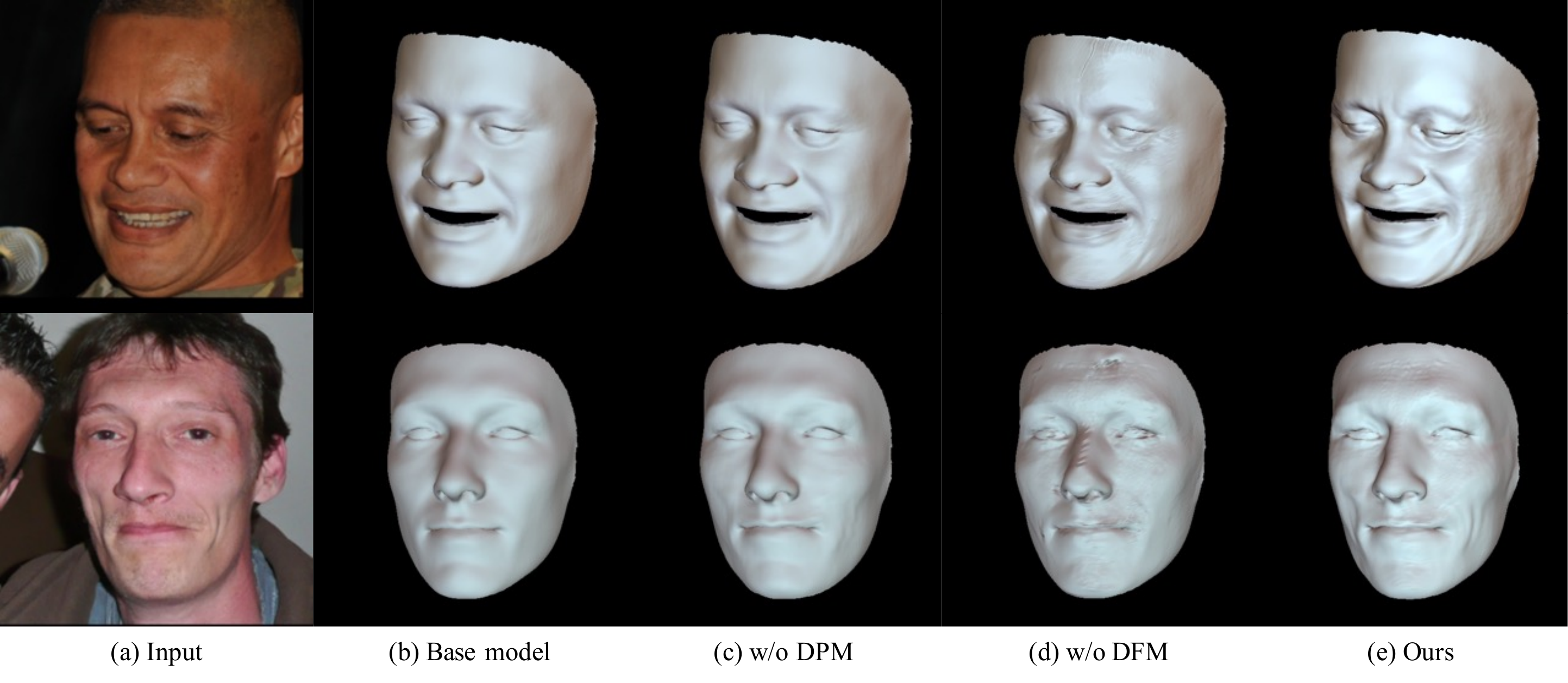} }
     \vspace{-5pt}
  \caption{Ablation study toward hierarchical modeling on FFHQ.}
  \label{fig: ablation_hierarchical_modeling}
  \vspace{-10pt}
\end{figure} 

\begin{figure}[t]
  \centering
  \resizebox{0.98\linewidth}{!}{
   \includegraphics{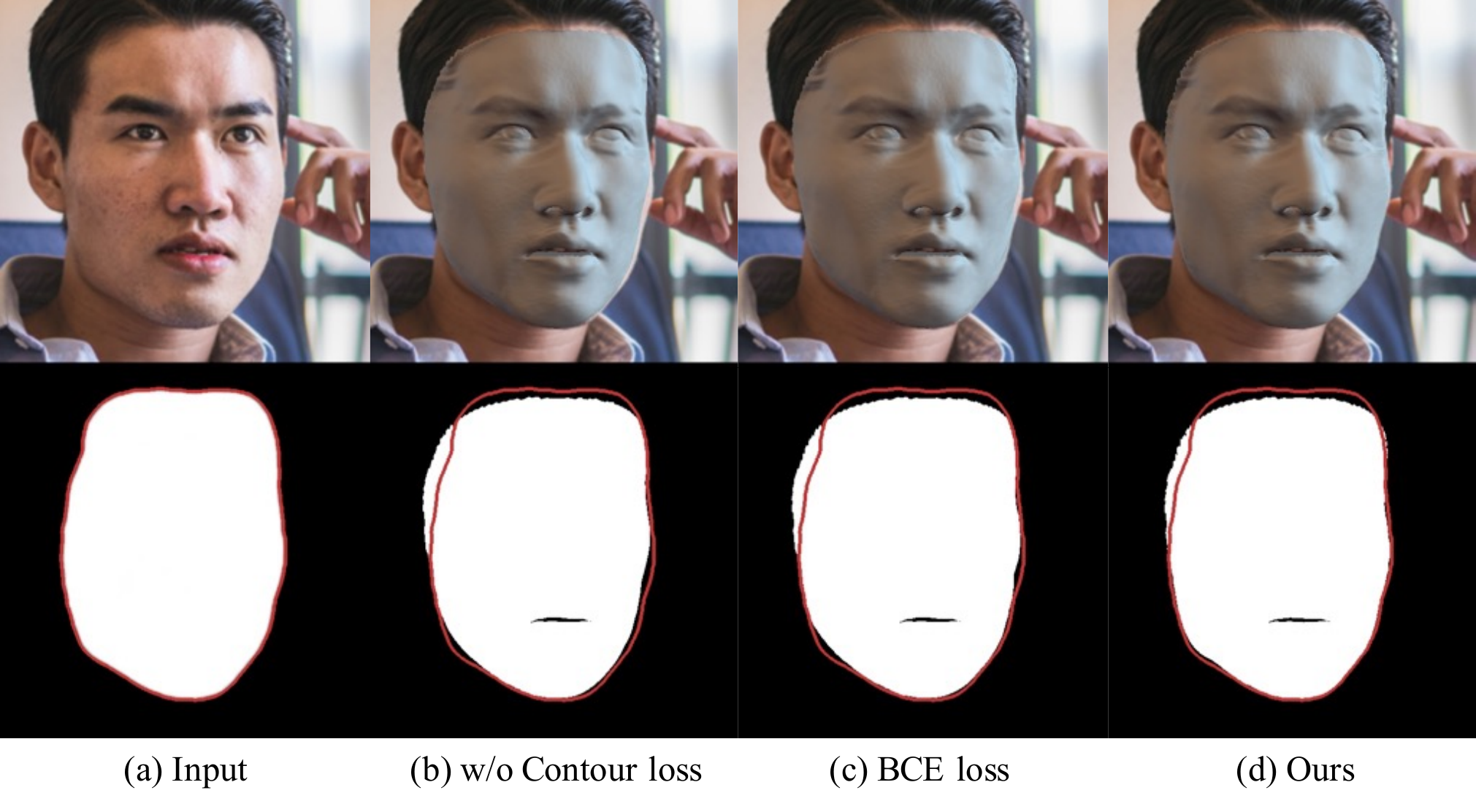} }
     \vspace{-5pt}
  \caption{Ablation study toward contour loss on FFHQ.}
  \label{fig: ablation_contour_loss}
  \vspace{-5pt}
\end{figure} 

\begin{figure}[t]
  \centering
  \resizebox{0.98\linewidth}{!}{
   \includegraphics{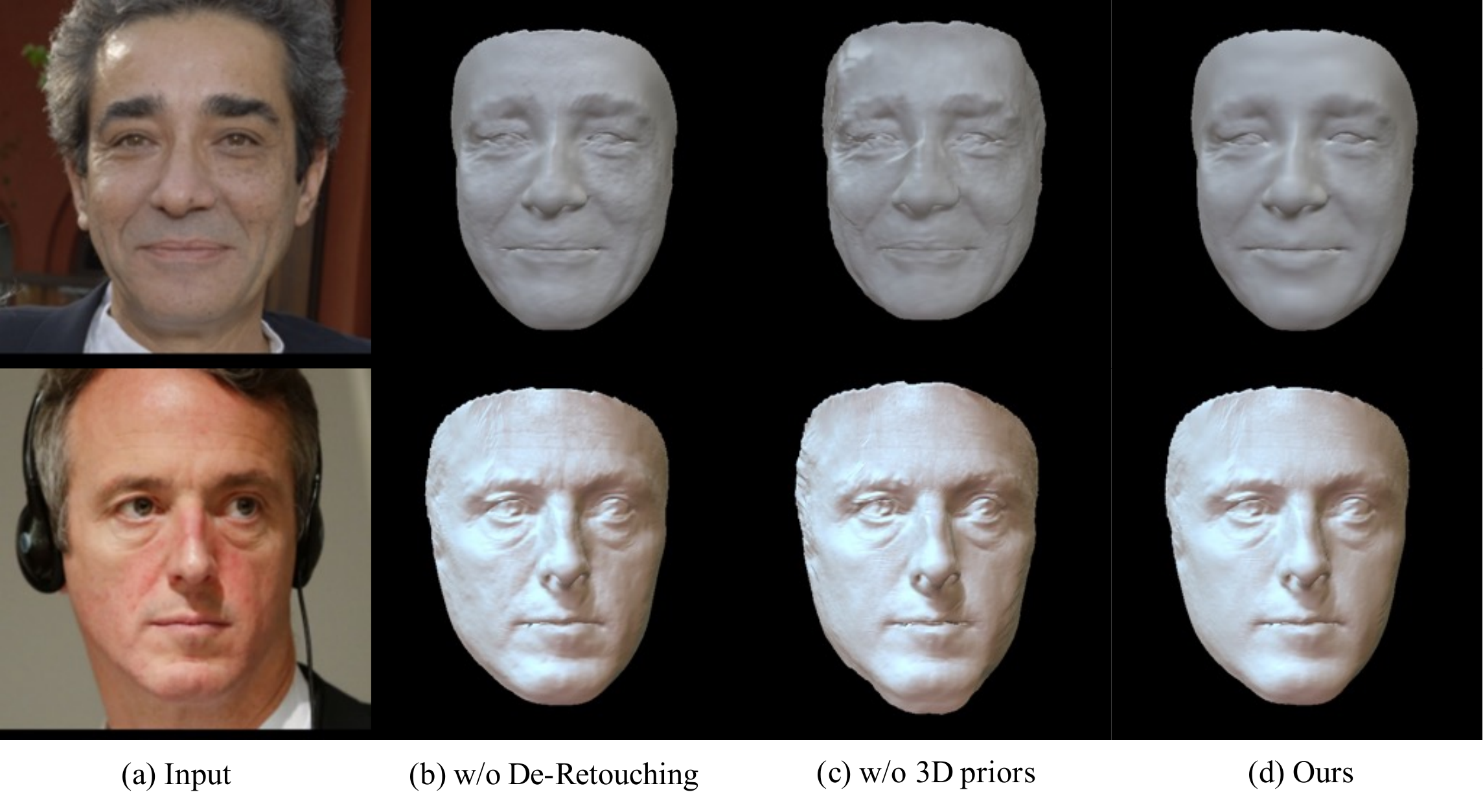} }
     \vspace{-5pt}
  \caption{Ablation study toward de-retouching module and 3D detail priors on FFHQ.}
  \label{fig: ablation_de_priors}
  \vspace{-5pt}
\end{figure} 

\begin{table}[t]\begin{center}
 \setlength{\belowcaptionskip}{-0.5cm}
 \setlength{\abovecaptionskip}{0.1cm}
% \resizebox{85mm}{12mm}{
\resizebox{1.00\linewidth}{!}{
\begin{tabular}{ccccc|c|c}
\toprule
Base model & HRM & $L_{con}$ & detail priors & DRM & CD(mm) & MNE(rad) \\ \hline
\checkmark &  &  &  &  & 3.8 & 0.092 \\
\checkmark & \checkmark &  &  &  & 3.34 & 0.081 \\
\checkmark & \checkmark & \checkmark &  &  & 3.18 & 0.079 \\
\checkmark & \checkmark & \checkmark & \checkmark &  & \textbf{2.9} & 0.067 \\
\checkmark & \checkmark & \checkmark & \checkmark & \checkmark & 2.91 & \textbf{0.065} \\
\bottomrule
\end{tabular}}
\caption{Quantitative ablation experiments on FaceScape-wild. HRM denotes the hierarchical representation modeling.}\label{table:ablation}
\end{center}
\end{table}

\begin{table}[t]\begin{center}
 \setlength{\belowcaptionskip}{-0.5cm}
 \setlength{\abovecaptionskip}{0.1cm}
\resizebox{1.00\linewidth}{!}{
\begin{tabular}{c|cccc|c}
\toprule
Methods & 2 views & 3 views & 4 views & 5 views & naive version (5 views) \\ \hline
Ours & 1.17 & 1.13 & 1.11 & \textbf{1.10} & 1.23 \\
\bottomrule
\end{tabular}}
\caption{Quantitative ablation study toward sparse-view reconstruction on FaceScape. Only median distance(mm) is reported.}\label{table:ablation_MV-HRN}
\end{center}
\end{table}

In order to verify the rationality and effectiveness of our network design, we conduct extensive ablation experiments on FFHQ and FaceScape. Table \ref{table:ablation} shows the quantitative results on FaceScape-wild benchmark. As revealed in the table, compared to the base 3DMM, the hierarchical modeling strategy brings a huge improvement ($\sim$0.5mm). The contour loss produces 0.16mm improvement. 3D priors of details play a key role in our framework, achieving $\sim$0.3mm improvement. The quantitative contribution of the de-retouching module is minor, while the following qualitative results on FFHQ prove its effectiveness.

\noindent {\bf On hierarchical modeling.} To demonstrate the necessity of the hierarchical modeling, we employ the deformation map (DFM) and displacement map (DPM) respectively (columns c, d in Fig.~\ref{fig: ablation_hierarchical_modeling}) to solely learn the overall facial details and compare the results. Without DPM, the DFM exhibit the capability of capturing high-frequency details to a certain extent. However, the performance is limited by the mesh density and the trade-off between the MF details and HF details. In contrast, the DPM is more effective in learning some micro details, but it fails to handle some larger-scale deformation. Apparently, by introducing the hierarchical modeling strategy, the proposed method achieves more accurate and detailed reconstruction.

\noindent {\bf On contour-aware loss.} The contour-aware loss $L_{con}$ aims to enhance the reconstruction accuracy of the facial contours. As shown in Fig.~\ref{fig: ablation_contour_loss}, the network with $L_{con}$ exhibit superior performance on learning the face contour compared with the one without $L_{con}$ or with BCE loss.

\noindent {\bf On 3D priors of facial details.} The introduced 3D priors provide the geometry distribution of the real facial details, which guide the model to achieve accurate reconstructions. As shown in Fig.~\ref{fig: ablation_de_priors}, the network without 3D priors produces some unrealistic details. In contrast, the reconstruction results with 3D priors are smoother and more faithful.

\noindent {\bf On de-retouching module.} The de-retouching module is proposed to achieve better decoupling of the facial appearance and geometry. As shown in Fig.~\ref{fig: ablation_de_priors}, without DRM, the model is more susceptible to the distraction of various skin textures and yields some nonexistent details. 

\noindent {\bf On sparse-view reconstruction.} We test the performance of MV-HRN on FaceScape given different numbers of input views. As Shown in Table \ref{table:ablation_MV-HRN}, MV-HRN achieves comparable results given only two views for input. With the increase of the number of views, the performance is gradually better, showing the ability of MV-HRN to aggregate multi-view information. Besides, we compare MV-HRN with its naive version (simply average the results of all views) and the results demonstrate the effectiveness of the network design.

\subsection{Extention work and Limitation}
Due to the limited paper space, more extension work (such as high-quality head reconstruction) and limitation of our method are provided in the supplementary materials.

\section{Conclusion}
In this paper, we propose a novel hierarchical representation network(HRN) for accurate and detailed face reconstruction from in-the-wild images. Specifically, we achieve facial geometry disentanglement and modeling by hierarchical representation learning. The 3D priors of details are further incorporated to improve the reconstruction results in accuracy and visual effects.
% we implement geometry disentanglement and introduce the hierarchical representation to fulfill detailed face modeling. Meanwhile, 3D priors of facial details are incorporated to enhance the accuracy and authenticity of the reconstruction results. 
Besides, we propose a de-retouching network, which alleviates the ambiguities between geometry and appearance. Moreover, we extend HRN to a multi-view fashion and introduce a high-quality 3D face dataset FaceHD-100 to boost the research of sparse-view FR. Extensive experiments reveal that our method achieves superior performance to the existing methods in terms of accuracy and visual effects.

\clearpage

%%%%%%%%% REFERENCES
{\small
\bibliographystyle{ieee_fullname}
\bibliography{egbib}
}

\end{document}